\documentclass{article}
\usepackage{spconf,amsmath,graphicx}

\usepackage{enumitem}
\setlist{nosep, leftmargin=14pt}

\usepackage{mwe} 

\title{Path-GPTOmic: A Balanced Multi-modal Learning Framework \\ 
for Survival Outcome Prediction}
\name{{\textit{Hongxiao Wang}$^{1}$ \qquad \textit{Yang Yang}$^{2}$ \qquad \textit{Zhuo Zhao}$^{3}$ \qquad Pengfei Gu$^{3}$\qquad \textit{Nishchal Sapkota}$^{3}$ \qquad \textit{Danny Z. Chen}$^{3}$ }}
\address{
$^{1}$ Capital Normal University, Beijing, 100048, China \\
$^{2}$ Novartis, San Diego, CA 92121, USA \\
$^{3}$ University of Notre Dame, Notre Dame, IN 46556, USA }
\begin{document}
%
\maketitle
\begin{abstract}
For predicting cancer survival outcomes, standard approaches in clinical research are often based on two main modalities: pathology images for observing cell morphology features, and genomic (e.g., bulk RNA-seq) for quantifying gene expressions. However, existing pathology-genomic multi-modal algorithms face significant challenges: (1) Valuable biological insights regarding genes and gene-gene interactions are frequently overlooked; (2) one modality often dominates the optimization process, causing inadequate training for the other modality.
In this paper, we introduce a new multi-modal ``Path-GPTOmic" framework for cancer survival outcome prediction. First, to extract valuable biological insights, we regulate the embedding space of a foundation model, scGPT, initially trained on single-cell RNA-seq data, making it adaptable for bulk RNA-seq data. Second, to address the imbalance-between-modalities problem, we propose a gradient modulation mechanism tailored to the Cox partial likelihood loss for survival prediction. The contributions of the modalities are dynamically monitored and adjusted during the training process, encouraging that both modalities are sufficiently trained. Evaluated on two TCGA(The Cancer Genome Atlas) datasets, our model achieves substantially improved survival prediction accuracy.

\end{abstract}
\section{Introduction}
\label{sec:intro}
Pathology images and genomic assays are two main data sources for predicting the survival outcome of cancer patients. Pathology images contain information on cell mitosis, cell morphology, and the micro-environment. Genomic data from RNA-seq measure the abundance of RNA transcripts, providing critical biological insights into cell identity, cellular activity, stage of development and differentiation, as well as cell functionality~\cite{cui2023scgpt}. In clinical practice, bulk RNA-seq (average global gene expression among cells) is more cost-effective, and is widely employed for analyzing cancer initiation and progression, as well as predicting survival outcomes~\cite{li2021bulk}. 


To jointly utilize these two complementary sources for automatic cancer survival prediction, researchers have developed a series of multi-modality deep learning algorithms. For example, in~\cite{chen2020pathomic,chen2022pan}, genomic data were processed with self-normalizing networks (SNN) and fused with pathology image embeddings with Kronecker product.  Transformer models were employed to capture genotype-phenotype interactions through an attention mechanism~\cite{chen2021multimodal}.  Different modalities were projected into the same latent space to enclose distances between multi-modal embeddings of the same patients~\cite{ding2023pathology,cheerla2019deep}. 

However, these methods still face two major challenges. First, the biological insight of genomic data cannot be fully explored by SNN or Transformer models with limited training data, and external knowledge on human cells is not utilized to compute more accurate genomic embeddings. Second, researchers found that the dominant modality (the one with better performance) may suppress the training process of the other modality~\cite{peng2022balanced, xu2023mmcosine}. Consequently, the other modality may not generalize well to test data.

In this paper, we propose a new multi-modal Path-GPTOmic framework, which combines pathology images and genomic data from patient specimen for predicting cancer survival outcomes.
To address the first challenge, we seek to learn a smooth latent space for bulk RNA-seq embeddings by incorporating a foundation model, scGPT~\cite{cui2023scgpt}. This model was originally trained on single-cell RNA-seq data from large human cell atlases~\cite{regev2017science}, and has demonstrated superior performance in cell functionality analysis (e.g., cell type annotation and genetic perturbation prediction). However, in our prior experiments, we observed that directly applying scGPT to bulk RNA-seq improved model performance only marginally for downstream tasks. One possible reason for this is that the direct application maps bulk RNA to an unfair latent space, where distances between embedding vectors do not accurately reflect RNA similarities.
To address this issue, we adopt generative model practices~\cite{liu2022smooth, wang2020unlabeled, beckham2019adversarial} to smooth the latent space with mix-up regulation. Specifically, we append a smoothing module (i.e., a multi-layer-proceptron (MLP) network) after scGPT. Then we train it by simulating bulk RNA-seq with individual single-cell RNA expression~\cite{li2021bulk}. This approach enables us to achieve an interpolatable latent space for gene expression, enhancing generalizability for bulk RNA data.

To tackle the second challenge, we combine the genomic and image branches of the model and closely monitor the contribution of each modality in training process. We find that genomic branch outperforms and contributes less to the Cox partial likelihood loss~\cite{ching2018cox}. As a result, the genomic branch dominates the training process, causing under-optimization of the image branch. To alleviate this issue in survival outcome prediction, we propose to control the loss optimization process by dynamically adjusting the gradient. Specifically, we assess the contributions of the two branches and appropriately modulate the gradient of the under-optimized image modality.

Our contributions can be summarized in three key aspects. (1) In clinical research, bulk RNA-seq data are more cost-effective and easier to acquire. To our best knowledge, we are the first to extend the single-cell foundation model scGPT for processing bulk RNA-seq data of patient samples. (2) We take pioneering steps to address the training imbalance problem in multi-modal pathology-genomics fusion tasks. (3) We evaluate our new method using two TCGA datasets, and obtain performance improvement compared to the baselines.

\section{Method}
\label{sec:method}
\begin{figure*}[htb]
\centering
  \centerline{\includegraphics[width=14cm,height=6.5cm]{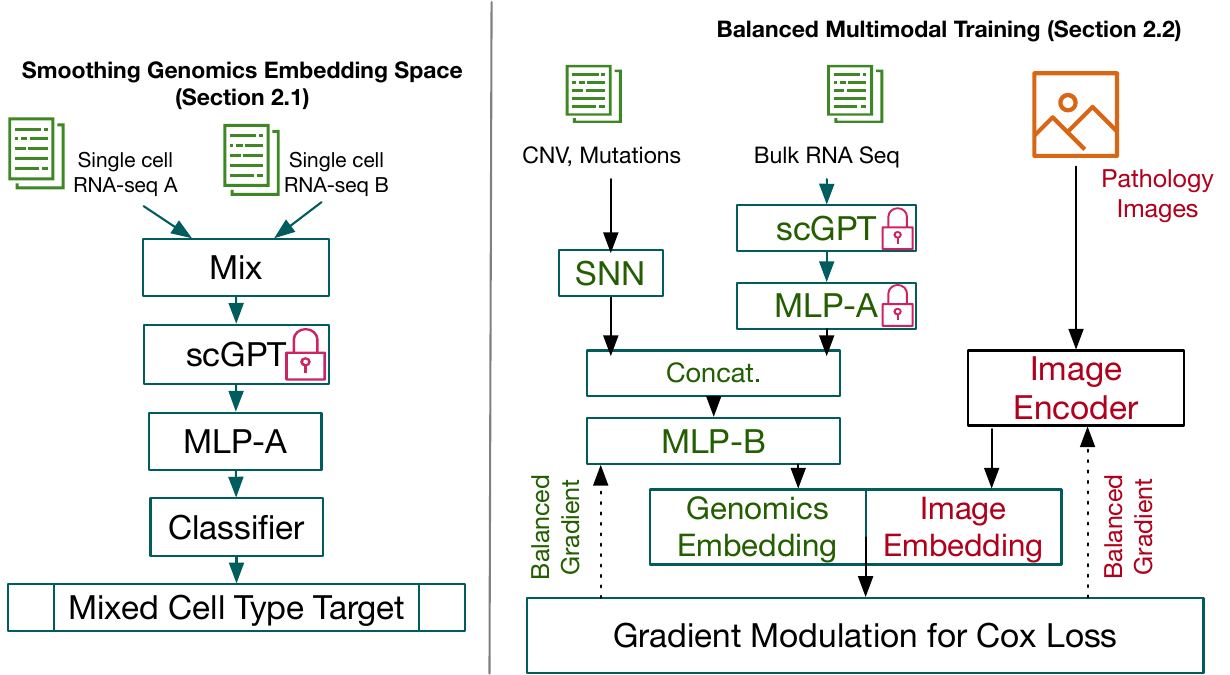}}
\caption{Illustrating our Path-GPTOmic pipeline. First, we train MLP-A for regulating the latent space for bulk RNA-seq embeddings, freezing the scGPT parameters. Second, we train SNN and MLP-B for genomics embedding and Image Encoder (T2T-ViT~\cite{yuan2021tokens}) for embedding pathology image features, freezing both the scGPT parameters and MLP-A parameters. }
\label{fig:framework}
\end{figure*}

Fig.~\ref{fig:framework} gives an overview of our method. In Section~\ref{sec:m1}, we show how to smooth the scGPT-derived single-cell RNA-seq embedding space, enabling scGPT to be adapted for bulk RNA-seq embeddings. In Section~\ref{sec:m2}, we describe how to balance the contributions of the genomics branch and image branch in training Cox partial likelihood loss, encouraging better optimization for both modalities. 

\subsection{Regulating the Genomics Embedding Space}
\label{sec:m1}
Cancer survival outcome predictions often require profiling transcript abundance through RNA-seq, including both single-cell RNA-seq (scRNAseq) and bulk RNA-seq (average global gene expression). On one hand, scRNAseq can provide fine-grained individual cell-level evidence for cancer. By training a Transformer model on large datasets, scGPT has uncovered gene-gene interactions and achieved superior performance on various downstream tasks. On the other hand, bulk RNA is more cost-effective and widely used in clinical research. Hence, it is desirable to extend scGPT to the analysis of bulk RNA-seq data for survival outcome prediction.

Our preliminary results show that directly applying scGPT to bulk RNA data is not significantly beneficial for downstream tasks (see Section~\ref{sec:expriment}). One possible reason is that without sufficient average single-cell RNA-seq input for training, the intermediate interpolation area of the latent space is abrupt and does not generalize well for global average data (i.e., bulk RNA-seq). A similar phenomenon has been observed in generation tasks. For example, as shown in~\cite{liu2021smoothing, peychev2022latent}, a model $f(\cdot)$ performing well on inputs $x_1, x_2$ may collapse and yield artifact results for interpolated input $(\lambda x_1 + (1-\lambda) x_2)$, with $\lambda \in [0,1]$.

By applying scGPT to bulk RNA-seq, our goal is to regulate the latent space using Mixup-based regulation~\cite{liu2022smooth, berthelot2019mixmatch}, encouraging ``average'' RNA-seq input to yield reasonable embeddings that are beneficial for downstream tasks. We choose cell type annotation~\cite{cui2023scgpt} prediction as the probing task, as it is closely related to survival outcomes. Our pipeline is outlined in the top part of Fig.~\ref{fig:framework}. Two randomly selected scRNAseq samples are interpolated by a mix layer, processed by a pre-trained scGPT module, a multi-layer perceptron (MLP) network (called as MLP-A), and a linear layer as a classifier. By encouraging the model output to converge to average cell types, we expect the embedding space (i.e., the output of MLP-A) to be regulated for representing average cell information. 


Our model is pre-trained as follows. First, in each optimization step, we randomly select two scRNAseq samples from the scGPT assembled human cell scRNAseq dataset. Formally, we denote the input scRNAseq as $x_i, x_j$, represented as one-hot cell type label encodings $\bf{t_i}, \bf{t_j}$, respectively. Second, we simulate the bulk RNAseq by interpolating $x_i, x_j$ with $(\lambda x_i + (1-\lambda) x_j)$, where $\lambda$ is a scalar value uniformly and randomly chosen from $[0,1]$. Third, to avoid excessively perturbing the well-trained scGPT model, we fix the scGPT parameters and append a three-layer MLP-A to it to regulate the scGPT output. Then, we optimize the output with a regression target value $(\lambda \bf{t_i} + (1-\lambda) \bf{t_j})$. In this way, the distance between the latent spaces can represent the weighted contributions of the RNAseq from the two input samples.

\subsection{Balanced Multi-modal Training}
\label{sec:m2}
In the optimization process of multi-modal deep learning models, researchers have observed that the dominant model, which displays better performance, tends to suppress the optimization of the others~\cite{peng2022balanced, xu2023mmcosine}. While the issue of imbalanced training has been addressed in optimizing cross-entropy loss, it has not been thoroughly investigated in the context of pathology-genomics multi-modal fusion for optimizing Cox partial likelihood loss in survival prediction.

\noindent\textbf{Framework.}
To better discuss the multi-modal imbalance training problem, we use a common prototype model, as shown in Fig.~\ref{fig:framework}. In this model, the Image Encoder submodule can be flexibly replaced with alternatives such as ResNet~\cite{chen2020pathomic}, Transformer~\cite{ding2023pathology}, or T2T-ViT~\cite{yuan2021tokens}. We define the training set as $\mathcal{D} = \{(g_k, p_k)\}_{k=1}^K$, where $K$ is the number of paired genomic data $g_k$ and pathology images $p_k$. The genomic data consist of copy number variations (CNV), mutations, and RNA-seq. We process the CNV and mutations with SNN in the same structure as in~\cite{chen2020pathomic}, yielding a vector $\bf{G^{(1)}_k}$. The RNA-seq is processed by scGPT and MLP-A sequentially, with the network parameters pre-trained and fixed as described in Section~\ref{sec:m1}, yielding a vector $\bf{G^{(2)}_k}$. These are then concatenated, and we train a new MLP-B to produce a genomic feature $\bf{G_k}=MLP_B([\bf{G^{(1)}_k}||\bf{G^{(2)}_k}])$. In parallel, the pathology image $p_k$ is processed by the Image Encoder to obtain an image feature $\bf{P_k}$. We then concatenate $\bf{G_k}$ and $\bf{P_k}$ as $[\bf{G_k}||\bf{P_k}]$. The log hazard ratio $\theta_k$ for patient $k$ is derived by using a linear classifier with trainable weights $W$ and bias $b$, as $\theta_k = W([\bf{G_k}||\bf{P_k}]) + b$.

Following~\cite{peng2022balanced}, we represent $W$ as a combination of two blocks $\it{W^G}$ and $\it{W^P}$, and rewrite the equation as:
\begin{equation}
    \theta_k = \it{W^G} \cdot \bf{G_k}  +  \it{W^P} \cdot \bf{P_k} + b.
\end{equation}

To predict survival outcome, Cox partial log-likelihood~\cite{ching2018cox,chen2020pathomic} is used as cost function:
\begin{equation}
    L_{Cox} =  \sum_{C(k)=1} (\theta_k - \log \sum_{R(t_k)} \exp({\theta_k})),
\end{equation}
where $C(k)=1$ represents the uncensored events, 
and $R(t_k)$ is the risk set at time $t_k$. The gradient is:
\begin{align}
        \frac{\partial L_{Cox}}{\partial \theta_k} &= \sum_{C(k)=1} (1 - \frac{\exp({\theta_k})}{\sum_{R(t_k)}\exp({\theta_k})}) \\
        &= \sum_{C(k)=1} (1 - \frac{e^{({W^G G_k + W^P P_k + b})}}{\sum_{R(t_k)}e^{({W^G  G_k + W^P P_k + b})}}).
\end{align}
\label{eq:grad}

One can observe that in the above equation, when one modality (say, the genomics modality) performs well, it will dominate the gradient loss in Equation (4) via $\it{W^G} \cdot \bf{G_k}$. The global loss will be close to zero. Thus, limited training optimization will be applied to the image modality. Even when the model converges, the image modality possibly remains inadequately trained.

\noindent\textbf{Balanced Training}. 
To alleviate the imbalance training problem, we evaluate the contribution discrepancy ratio $\rho^G, \rho^P$ for Cox partial log-likelihood loss, as:
\begin{align}
    \rho^G &= ( \frac{({W^G  \bf{G_k} + \frac{b}{2}})}{\sum_{R(t_k)}e^{({W^G  \bf{G_k} + \frac{b}{2}})}})/( \frac{({W^P  \bf{P_k} + \frac{b}{2}})}{\sum_{R(t_k)}e^{({W^P  \bf{P_k} + \frac{b}{2}})}}), \\ 
    \rho^P &= 1/\rho^G.
\end{align}
Formally, inspired by~\cite{peng2022balanced}, we modulate the gradient by estimating the contributions of the two modalities. We update the model parameters in each iteration $t$, as:
\begin{align}
    \phi^G_{t+1} &= \phi^G_{t} - \eta \cdot \min (1-\text{tanh}(\rho^G-1),1) \cdot g(\phi^G_t), \\
    \phi^P_{t+1} &= \phi^P_{t} - \eta \cdot \min (1-\text{tanh}(\rho^P-1),1) \cdot g(\phi^P_t),
\end{align}
where $\phi^G_{t+1}$ and $g(\phi^G_t)$ represent the parameters and gradient for the MLP in the genomics network, $\phi^P_{t+1}$ and $g(\phi^P_t)$ represent the parameters and gradient for the pathology image network, and $\eta$ is the learning rate. In this way, the learning rate of the modality with higher $\rho$ is suppressed.

\section{Experiments}
\label{sec:expriment}

\subsection{Datasets and Implementation Details}
\label{sec:expdetail}
Our experiments utilize two datasets~\cite{chen2020pathomic} that consist of hematoxylin and eosin (H\&E)-stained pathology images, corresponding genomic  features (mutations, copy number variations (CNV), RNA-seq), and patient survival outcomes. Specifically, the TCGA-GBMLGG dataset contains 1505 gliomas (brain and spinal cord tumors) samples, and the TCGA-KIRC dataset contains 1251 clear cell renal cell carcinoma samples. We apply the same experimental protocol as~\cite{chen2020pathomic} by evaluating model performance with 15-fold cross-validation. 

In our pipeline, the CNV and mutation information is handled by SNN with the same architecture as~\cite{chen2020pathomic}. The bulk RNA-seq data are processed with scGPT~\cite{cui2023scgpt}. Both MLP-A and MLP-B are three-layer perceptron networks with hidden layer dimension 128. The classifier is a linear layer mapping the MLP-A generated feature vector to 17 cell type categories. The batch size is 32. Our model is implemented with PyTorch and is trained on an NVIDIA A10 GPU. 

\subsection{Results}
\noindent\textbf{Main Results.} Table~\ref{table:expgbmlgg} and Table~\ref{table:expkirc} show our main results. Our model is compared with the baselines of SCNN, SGCNN~\cite{mobadersany2018predicting}, Pathomic Fusion~\cite{chen2020pathomic}, and the supervised multi-modal setting in~\cite{ding2023pathology}. For fair comparison, we also replace the CNN in Pathomic Fusion~\cite{chen2020pathomic} with the T2T-ViT~\cite{yuan2021tokens} backbone to extract features from pathology images. We use C-Index to measure performance. First, our model outperforms all the baselines on both datasets. Second, similar to the setting ``Pathomic Fusion~\cite{chen2020pathomic} (T2T-ViT~\cite{yuan2021tokens} + SNN)'', we also use the backbone T2T-ViT and SNN to process the image features and CNV features. Our model gains around $2\%$ improvement on both datasets. This indicates the effectiveness of our design for regulating scGPT embeddings and gradient modulation. 

\begin{table}[htb]
\centering
\caption{Comparison of C-Index performance on the TCGA-GBMLGG dataset ($p<0.05$). }
\begin{tabular}{l|l}
\hline
              Method                                 & C-Index \\\hline
SCNN (Histology Only)~\cite{mobadersany2018predicting}                          & 0.754  \\\hline
Histology CNN~\cite{chen2020pathomic}                                  & 0.792 ± 0.014  \\\hline
Histology T2T-ViT~\cite{yuan2021tokens}                                 & 0.803 ± 0.016  \\\hline
Genomic SNN~\cite{chen2020pathomic}                                   & 0.808 ± 0.014  \\\hline \hline
GSCNN (Histology + Genomic)~\cite{mobadersany2018predicting}                 & 0.781    \\\hline
Pathomic Fusion~\cite{chen2020pathomic} (CNN + SNN)                   & 0.820 ± 0.009  \\\hline
Pathomic Fusion~\cite{chen2020pathomic} (T2T-ViT~\cite{yuan2021tokens} + SNN)                   & 0.826 ± 0.010  \\\hline
PathOmics~\cite{chen2022pan}                      & 0.833 ± 0.012  \\\hline
Ours  & \textbf{0.848 ± 0.014}   \\ \hline
\end{tabular}
\label{table:expgbmlgg}
\end{table}

\begin{table}[htb]
\centering
\caption{Comparison of C-Index performance on the TCGA-KIRC dataset ($p<0.05$).}
\begin{tabular}{l|l}
\hline
         Method                                      & C-Index \\\hline
Histology CNN~\cite{chen2020pathomic}                                  & 0.671 ± 0.023   \\\hline
Histology T2T-ViT~\cite{yuan2021tokens}                                 & 0.683 ± 0.023    \\\hline
Genomic SNN~\cite{chen2020pathomic}                                   & 0.684 ± 0.025   \\\hline \hline
Pathomic Fusion~\cite{chen2020pathomic} (CNN + SNN)                   & 0.719 ± 0.031   \\\hline
Pathomic Fusion~\cite{chen2020pathomic} (T2T-ViT~\cite{yuan2021tokens} + SNN)                   & 0.727 ± 0.033   \\\hline
PathOmics~\cite{chen2022pan}                      & 0.736 ± 0.024   \\\hline
Ours  & \textbf{0.754 ± 0.030}   \\ \hline
\end{tabular}
\label{table:expkirc}
\end{table}

\noindent\textbf{Ablation Study.} We evaluate the impact of each of our key model components in Table~\ref{table:expablation}. First, we find that directly applying scGPT is a sub-optimal solution as the performance is improved only by $0.05\%$ (see Exp.~1 and Exp.~2). Second, by incorporating our mix-up regulation module to smooth the latent space, the scGPT's ability to process bulk RNA genomics data is realized (see Exp.~1 and Exp.~4). Third, our gradient modulation (GradMod.) can effectively help improve the performance, in comparison with direct concatenation and the performance with the Kronecker product-based fusion in~\cite{chen2020pathomic} (see the comparison between Exp.~2 and Exp.~3, and the comparison among Exp.~4, Exp.~5, and Exp.~6).

\begin{table}[htb]
\centering
\caption{Ablation study on the TCGA-GBMLGG dataset.}
\begin{tabular}{l|l|l|l}
\hline
Exp. & Encoding RNA-seq   & Fusion                  & C-Index       \\ \hline
1      & SNN              & Concat                  & 0.826 ± 0.010 \\ \hline
2      & scGPT w/o smooth & Concat                  & 0.831 ± 0.012 \\ \hline
3      & scGPT w/o smooth & GradMod. & 0.840 ± 0.013 \\ \hline
4      & scGPT w/ smooth  & Concat                  & 0.838 ± 0.013  \\ \hline
5      & scGPT w/ smooth  & Kro. Prod.~\cite{chen2020pathomic}    & 0.840 ± 0.013  \\ \hline
6      & scGPT w/ smooth  & GradMod. & 0.848 ± 0.014 \\ \hline
\end{tabular}
\label{table:expablation}
\end{table}


\section{Conclusions}
\label{sec:conclusion}
In this paper, we addressed two main challenges in pathology image and genomics data fusion. First, we showed how to effectively use the advanced foundation model scGPT, originally designed for single cell RNA-seq, for processing bulk RNA-seq data. Second, we tackled the imbalance training problem between image modality and genomics modality by proposing gradient modulation for the Cox partial likelihood loss. Evaluated on two TCGA datasets, our model effectively improved the performance compared to the baseline models.

%

\bibliographystyle{IEEEbib}
\bibliography{strings,refs}

\end{document}